\documentclass[11pt]{article}

\usepackage{microtype} % microtypography
\usepackage{booktabs}  % tables
\usepackage{wrapfig}
\usepackage{amsmath}
\usepackage{amsthm}
\DeclareMathOperator*{\argmax}{arg\,max}

\usepackage{graphicx}
\graphicspath{ {./figures/} }

\usepackage[final]{automl}

\usepackage{algorithm}
\usepackage{algorithmic}
\title{Towards Personalized Preprocessing Pipeline Search} 

\author[1]{\nameemail{Diego Martinez}{dmartinez05@tamu.edu}}
\author[2]{\nameemail{Daochen Zha}{daochen.zha@rice.edu}}
\author[1]{\nameemail{Qiaoyu Tan}{qytan@tamu.edu}}
\author[2]{\nameemail{Xia Hu}{xia.hu@rice.edu}}

\affil[1]{Texas A\&M University}
\affil[2]{Rice University}

\hypersetup{%
  pdfauthor={}, % will be reset to "Anonymous" unless the "final" package option is given
  pdftitle={},
  pdfsubject={},
  pdfkeywords={}
}

\begin{document}

\maketitle

\begin{abstract}

%  since different features may require different feature transformation strategies

Feature preprocessing, which transforms raw input features into numerical representations, is a crucial step in automated machine learning (AutoML) systems. However, the existing systems often have a very small search space for feature preprocessing with the same preprocessing pipeline applied to all the numerical features. This may result in sub-optimal performance since different datasets often have various feature characteristics, and features within a dataset may also have their own preprocessing preferences. To bridge this gap, we explore personalized preprocessing pipeline search, where the search algorithm is allowed to adopt a different preprocessing pipeline for each feature. This is a challenging task because the search space grows exponentially with more features. To tackle this challenge, we propose ClusterP3S, a novel framework for \underline{P}ersonalized \underline{P}reprocessing \underline{P}ipeline \underline{S}earch via \underline{Cluster}ing. The key idea is to learn feature clusters such that the search space can be significantly reduced by using the same preprocessing pipeline for the features within a cluster. To this end, we propose a hierarchical search strategy to jointly learn the clusters and search for the optimal pipelines, where the upper-level search optimizes the feature clustering to enable better pipelines built upon the clusters, and the lower-level search optimizes the pipeline given a specific cluster assignment. We instantiate this idea with a deep clustering network that is trained with reinforcement learning at the upper-level, and random search at the lower level. Experiments on benchmark classification datasets demonstrate the effectiveness of enabling feature-wise preprocessing pipeline search.
\end{abstract}

\section{Introduction}

Feature preprocessing plays a crucial role in building a machine learning (ML) pipeline~\cite{zha2023data,wang2020skewness,zha2019multi,garcia2016big,tan2022bring,li2022towards}. It transforms the raw input features into numerical representations through multiple preprocessing primitive steps, such as missing value completion, normalization, etc. Feature preprocessing is so important that around 50\% of the time is spent on data preprocessing in building an ML system, reported in a survey collected from practitioners~\cite{munson2012study}. Thus, modern automated machine learning (AutoML) systems have included various preprocessing primitives in building ML pipelines~\cite{drori2021alphad3m,heffetz2020deepline,le2020scaling,liu2020admm,rakotoarison2019automated,lai2021tods,zha2021autovideo}.

Despite the great successes of the existing AutoML systems, they often have a very small search space for the feature preprocessing~\cite{drori2021alphad3m,feurer2015efficient,olson2016tpot,akhtar2018oboe}. A common strategy is to perform fixed transformations for non-numerical features and search for preprocessing pipelines from a small search space for numerical features. For example, Auto-Sklearn~\cite{feurer2015efficient}, one of the most popular AutoML systems, adopts a constant imputer and a one-hot encoder for all the non-numerical features, and searches for the imputers and scalers for numerical features. It applies the same preprocessing primitives to all the numerical features so that there are only 21 possible combinations within its search space\footnote{\scriptsize \url{https://github.com/automl/auto-sklearn/blob/master/autosklearn/pipeline/components/data_preprocessing/}}. While more recent systems, such as AlphaD3M~\cite{drori2021alphad3m}, have introduced more preprocessing primitives to handle different data types, the possible preprocessing pipelines still follow very strict grammars, and the same pipelines are applied for all the numerical features.

Unfortunately, this simple design of the preprocessing search space may lead to sub-optimal performance since different features may require different pre-processing pipelines to achieve the best results. For example, for the choices of encoders, some numerical features may only have very few possible values so it could be better to encode them like categorical features. For feature normalization, different features may have very different value distributions such that they may require different scalers. Motivated by this, we propose to allow the search algorithm to adopt a specific preprocessing pipeline for each feature. Specifically, we investigate the following research question: \emph{Can we improve the performance by enabling feature-wise preprocessing pipeline search?}

This is a non-trivial task because of two major challenges. First, the search space will grow exponentially with more features. Specifically, let $|\mathcal{P}|$ be the number of possible preprocessing pipelines, and $D$ be the number of features. Then the search space is $|\mathcal{P}|^D$. Thus, we need an efficient strategy to explore the search space. Second, many of the preprocessing pipelines can be invalid. For example, a mean imputer will raise errors when applied to string values. We need to avoid frequently sampling invalid pipelines to discover valid pipelines more easily.

To tackle these challenges, we propose ClusterP3S, a novel framework for \underline{P}ersonalized \underline{P}reprocessing \underline{P}ipeline \underline{S}earch via \underline{Cluster}ing. To efficiently explore the search space, we learn feature clusters, where the same preprocessing pipeline will be adopted for the features within a cluster. In this way, we can significantly reduce the search, which also inherently encourages the search algorithm to hit valid pipelines. In particular, we propose a hierarchical search strategy to jointly learn the clusters and search for the optimal pipelines, where the upper-level search optimizes the feature clustering, and the lower-level search optimizes the pipeline given a specific cluster assignment. We instantiate this idea with a reinforced deep clustering network at the upper-level, and random search at the lower-level. Extensive experiments on benchmark real-world classification datasets suggest that enabling feature-wise preprocessing pipeline search can significantly improve performance. To summarize, we make the following major contributions.
\vspace{-5pt}

\begin{itemize}
    \item Identify the importance of enabling feature-wise preprocessing pipeline search, which is often overlooked by the existing AutoML systems.
    \vspace{-5pt}
    \item Propose ClusterP3S framework for personalized preprocessing pipeline search. It adopts a hierarchical search strategy to jointly learn the clusters and search for the optimal pipelines.
    \vspace{-5pt}
    \item Instantiate ClusterP3S with a deep clustering network trained with reinforcement learning and random search for the upper-level search and the lower-level search, respectively.
    \vspace{-5pt}
    \item Demonstrate the effectiveness of ClusterP3S on eight benchmark classification datasets, showing the promise of personalized preprocessing pipeline search. In addition, we also present ablation and hyperparameter studies to understand how ClusterP3S behaves.
    \vspace{-5pt}
\end{itemize}

\section{Problem Formulation}

Given a training dataset $\mathcal{D}_{\text{train}} = \{(\mathbf{x}_i, y_i)\}_{i=1}^{N}$, where $\mathbf{x}_{i} \in \mathbb{R}^D$ is a $D$-dimensional feature vector, and $y_i \in \mathbb{R}$ denotes the target, we aim to learn a mapping from features to targets $f: \mathbb{R}^D \to \mathbb{R}$ based on $\mathcal{D}_{\text{train}}$ such that $f$ can accurately make predictions on a validation dataset. Here, $f$ is often a pipeline consisting of multiple primitive steps, where a primitive is an implementation of a specific function and serves as a basic building block in a pipeline. In this work, we consider a typical pipeline design with multiple preprocessing steps, followed by a machine learning model. Additionally, we allow each feature to have a different preprocessing pipeline (which is a sub-pipeline of $f$). Formally, a pipeline $f$ can be represented as a $(D+1)$-tuple $\langle P_1, P_2, ..., P_D, M \rangle$, where $P_j$ ($j \in \{1,2,...,D\}$) is the preprocessing pipeline for the $j^\text{th}$ feature and can have multiple steps, and $M$ denotes a machine learning model that takes as input all the pre-processed features and outputs the predictions.

We describe the problem of \underline{P}ersonalized \underline{P}re-processing \underline{P}ipeline \underline{S}earch (P3S) as follows. Let $\mathcal{P}$ be the search space of the preprocessing pipelines. Given a training dataset $\mathcal{D}_{\text{train}}$, a validation dataset $\mathcal{D}_{\text{valid}}$, a machine learning model $M$, we aim to solve the following optimization problem:

\begin{equation}
\label{eq:1}
\argmax_{P_j \in \mathcal{P}} L (f, \mathcal{D}_\text{train}, \mathcal{D}_\text{valid}),
\end{equation}
where $f = \langle P_1, P_2, ..., P_D, M \rangle$ is the pipeline, and $L (f, \mathcal{D}_\text{train}, \mathcal{D}_\text{val})$ is the performance metric on $\mathcal{D}_\text{valid}$ when fitting $f$ on $\mathcal{D}_\text{train}$. In this work, we perform 10-fold cross validation to obtain $L$. Let $|\mathcal{P}|$ denote the number of possible preprocessing pipelines in $\mathcal{P}$. Then the search space is $|\mathcal{P}|^D$, which grows exponentially with more features. In our preliminary experiments, we find that naively applying an existing search algorithm to this massive search space will lead to poor performance.

\section{Methodology}

\begin{figure}[t]
\centering
\includegraphics[width=0.9\textwidth]{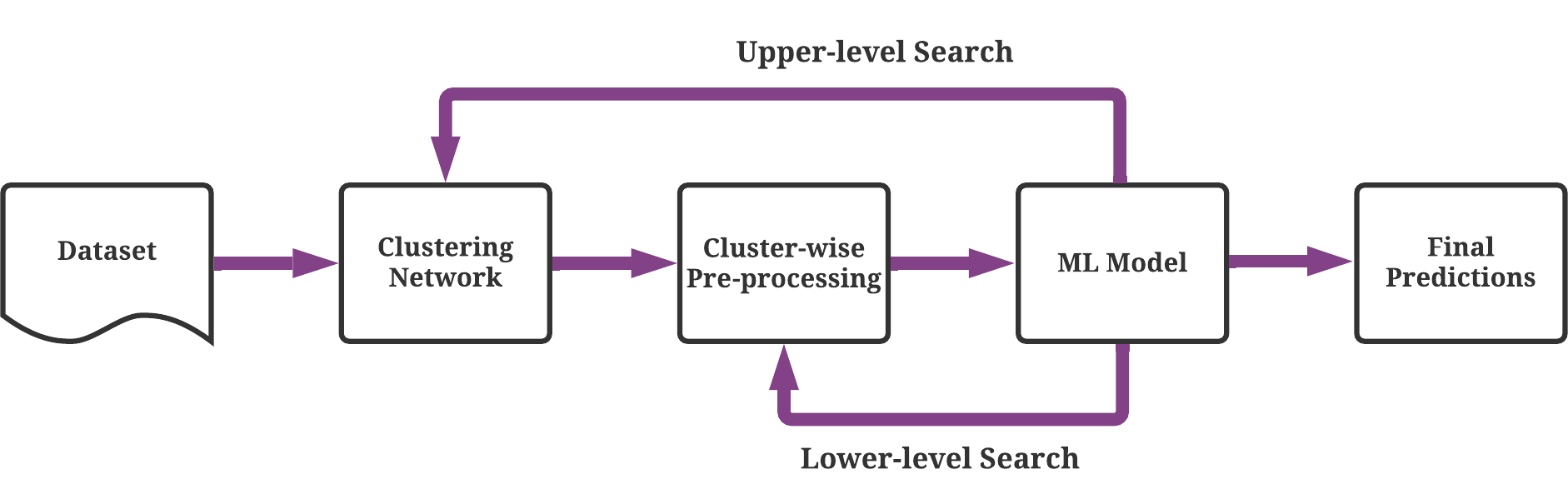}
\vspace{-8pt}
\caption{A high-level overview of the ClusterP3S framework. The column features are first embedded from the input dataset and then processed by a deep clustering network which assigns each feature to a cluster. The cluster information is forwarded to the cluster-wise pre-possessing module which applies a preprocessing pipeline to each cluster. Finally, an ML model will generate the predictions based on the pre-processed features. This process is jointly optimized following a hierarchical search strategy, where the upper-level search optimizes the feature clustering to enable better pipelines built upon the clusters, and the lower-level search optimizes the pipeline given a specific cluster assignment.}
\vspace{-8pt}
\label{fig:framework}
\end{figure}

%Clustered Feature-wise Pipeline Search Framework. An Autoencoder is used to generate the column feature embedded from the input dataset. Then a randomly initialized clustering network is used to generate the feature clusters. The cluster information is forwarded to the feature-wise random search that will eventually generate transformed data. Then the data is forwarded to the ML Learner to generate predictions used on the Policy Gradient Loss to generate a reward. Then, the reward is used to tune the Clustering Network. The process keeps repeating until no more resources are allocated.

We propose ClusterP3S for the P3S problem, illustrated in Figure~\ref{fig:framework}. ClusterP3S consists of (i) a deep clustering network which embeds the column features and assigns each feature to a cluster, (ii) a cluster-wise preprocessing module that samples a preprocessing pipeline for each cluster, and (iii) an ML model which generates predictions based on the pre-processed features. We first introduce a hierarchical search objective to jointly optimize the clustering network and the pipeline (Section~\ref{sec:method1}). Then we propose a piratical instance that achieves the upper-level search and the lower-level search with reinforcement learning and random search, respectively (Section~\ref{sec:method2}). Finally, we instantiate ClusterP3S with a tailored search space (Section~\ref{sec:method3}).

%High-level overview. Road-map
%Framework applies figure 1 upper lvl do something and lower do this.

%The ClusterP3S (Figure~\ref{fig:fwt_framework}) framework uses bi-level optimization strategy to jointly learn the clusters and search for the optimal pipelines, where the upper-level objective optimizes the feature clustering to enable better pipelines built upon the clusters, and the lower-level objective optimizes the pipeline given a specific cluster assignment. We instantiate this idea with a deep clustering network that is trained with reinforcement learning at the upper-level, and random search at the lower-level. 

\subsection{Hierarchical Search of Feature Clusters and Preprocessing Pipelines}
\label{sec:method1}

This subsection introduces a hierarchical search objective by clustering similar features and applying the same preprocessing pipeline to all the features within a cluster. Specifically, suppose the number of clusters is $K$. Then the search space can be reduced exponentially from $|\mathcal{P}|^{D}$ to $|\mathcal{P}|^{K}$.

A naive way to achieve this is to first embed the feature and then apply an off-the-shelf clustering algorithm, such as K-Means. However, this will lead to sub-optimal performance. First, the obtained clusters highly depend on the feature embedding, whose quality is hard to control. As a result, features that require different preprocessing pipelines could be falsely grouped together and forced to use the same pipeline. Second, the clusters obtained in this way are fixed in the whole search process. Bad clusters could significantly limit the performance upper bound in the search phase.

To tackle these issues, we propose to dynamically learn feature clusters in an end-to-end fashion. Let $\mathbf{c} \in \mathcal{C} \subseteq \mathbb{R}^{D}$ be the cluster assignment of the features, where each element $c_j \in \{1, 2, ..., K\}$, and $K$ is the number of clusters. We abuse the notations of $L$ and $f$ with a given cluster assignment by letting $L (f, \mathbf{c}, \mathcal{D}_\text{train}, \mathcal{D}_\text{valid})$ be the performance metric given $\mathbf{c}$, and $f = \langle P_1, P_2, ..., P_K, M \rangle$ which only searches for $K$ preprocessing pipelines. We aim to optimize the following objective:

\begin{equation}
\label{eq:2}
\argmax_{\mathbf{c} \in \mathcal{C}, P_k \in \mathcal{P}}  L (f, \mathbf{c}, \mathcal{D}_\text{train}, \mathcal{D}_\text{valid}).
\end{equation}
Eq~(\ref{eq:2}) can be interpreted as a hierarchical search problem with two levels: at the upper-level, we find the best clusters, and at the lower-level, we search for the best pipeline given a cluster assignment. While the overall search complexity of Eq~(\ref{eq:2}) is the same as that of Eq~(\ref{eq:1}), we can leverage the similarity pattern in the feature embedding space to effectively cluster the features such that we can reduce the upper-level search complexity significantly.

\subsection{Training Deep Clustering Network with Reinforcement Learning}
\label{sec:method2}
This subsection introduces a practical instance to achieve the hierarchical search based on a deep clustering network and reinforcement learning. Our design consists of four steps: (i) a feature embedding module using an AutoEncoder, (ii) a deep clustering network that assigns each feature to a cluster, (iii) a cluster-wise preprocessing pipeline search module for the lower-level search, and (iv) a reinforcement learning loss to update the clustering network.
%We will first elaborate on these fours steps and then summarize the overall learning procedure.

\noindent\textbf{Feature Embedding.} The goal of feature embedding is to generate a representation for each column feature. Motivated by how documents are encoded in text classification~\cite{korde2012text}, we treat each feature column as a document and the value of each instance within the column as a term. Then we use the term frequency of each column feature to represent the feature. Formally, let $V$ be the vocabulary size, i.e., the number of terms in the dataset. Each feature will be embedded as $\mathbf{e}_j \in \mathbb{R}^{V}$, where the $v^{\text{th}}$ element of $\mathbf{e}_j$ indicates the number of appearances of the $v^{\text{th}}$ term. Following this strategy, we can embed a dataset as $\mathbf{E} \in \mathbb{R}^{D \times V}$. However, $\mathbf{E}$ is often high-dimensional. Thus, we further use an AutoEncoder to obtain condensed embeddings. Specifically, we train an AutoEncoder with Mean-Squared-Error (MSE) loss:
\begin{equation}
\label{eq:3}
L_{\text{AE}} = (\text{AutoEncoder}(\mathbf{E}) - \mathbf{E})^2.
\end{equation}
Then we use the condensed representations obtained by the AutoEncoder as the final dataset embedding, denoted as $\widetilde{\mathbf{E}} \in \mathbb{R}^{D \times H}$, where $H$ is the hidden dimension of the AutoEncoder.

\noindent \textbf{Deep Clustering Network.} Given the dataset embedding $\widetilde{\mathbf{E}}$, we aim to assign each feature to a cluster. Motivated by~\cite{caron2018deep}, we use a deep clustering network to map feature embeddings to cluster IDs. The deep clustering network is initialized in an unsupervised fashion following two steps. First, we use a clustering algorithm, such as K-Means, to assign the features to $K$ clusters. Second, we generate pseudo labels based on the clusters and use supervised loss to train the clustering network. Formally, let $\text{ClusterNN}(\cdot)$ be the clustering network, and $\widetilde{\mathbf{c}}_j$ be the one-hot pseudo labels for the $j^{th}$ feature. We train the network with cross-entropy loss:

\begin{equation}
    L_{\text{cluster}} = \sum_{i=1}^N \text{cross-entropy}(\text{ClusterNN}(\widetilde{\mathbf{e}}_j), \widetilde{\mathbf{c}}_j),
    \label{eq:4}
\end{equation}
where $\widetilde{\mathbf{e}}_j$ is the embedding for the $j^{th}$ feature, and $\text{cross-entropy}(\cdot, \cdot)$ is the standard cross-entropy loss. However, the clusters obtained in this way may not be optimal. One benefit of using deep clustering instead of K-Means is that we can update the clustering network with gradient descent. Thus, Eq.~(\ref{eq:4}) is only used for initializing the clustering network, which will be further updated later. 

%Deep clustering formulation
%\begin{equation}
%\begin{aligned}
%&\min _{\mathcal{R}, \theta} \sum_{x \in \mathcal{X}} g(x, A(x ; \theta))+\lambda f\left(\mathbf{h}_{\theta}(x), c_{f}\left(\mathbf{h}_{\theta}(x) ; \mathcal{R}\right)\right) \\
%&\text { with: } c_{f}\left(\mathbf{h}_{\theta}(x) ; \mathcal{R}\right)=\underset{\mathbf{r} \in \mathcal{R}}{\operatorname{argmin}} f\left(\mathbf{h}_{\theta}(x), \mathbf{r}\right)
%\end{aligned}
%\end{equation}

\noindent \textbf{Cluster-wise Preprocessing Pipeline Search.} Given a clustering network $\text{ClusterNN}(\cdot)$, we can produce the cluster assignment for all the features with a forward pass. Formally, let $\mathbf{c} = \text{ClusterNN}(\widetilde{\mathbf{E}})$ be the cluster assignment, where $\mathbf{c} \in \mathbb{R}^{D}$, and its each element $c_j \in \{1,2,...,K\}$. Given $\mathbf{c}$, we search for the best preprocessing pipelines, i.e., $\argmax_{P_k \in \mathcal{P}}  L (f, \mathbf{c}, \mathcal{D}_\text{train}, \mathcal{D}_\text{valid})$. In this work, we adopt the random search, where in each iteration, we randomly sample a preprocessing pipeline. Since there can be many invalid pre-processing pipelines in the search space, we record the invalid primitives met in the search and force the algorithm not to sample the invalid ones later.

%The cluster-wise preprocessing pipeline search module will output the preprocessing pipeline that leads to the best performance in search.

% Intuitively, if we find that a cluster assignment always leads to poor performance, we should discourage the clustering network to generate such clusters. In contrast, if a clustering assignment tends to lead to strong performance, we should encourage this behavior. This trial-and-error reasoning aligns with reinforcement learning. Thus, we are motivated to 

\noindent \textbf{Reinforcement Learning Update.} The initial cluster assignment of the clustering network could be sub-optimal in that it is learned in an unsupervised way and it is not performance-aware. We propose to use reinforcement learning to update the clustering network towards the best cluster assignment. Specifically, we treat the clustering network as a policy network, whose outputs are actions indicating the probability of assigning the input feature to the $K$ clusters. Then we use reinforcement loss to update the policy network using the performance as the reward. Formally, let $p(c_j|\widetilde{\mathbf{e}}_j)$ be the probability of assigning the $j^{th}$ feature to the cluster $c_j$, $r_{\text{perf}}$ be the reward, i.e., the classification performance. To reduce the variance, we employ a baseline in the reward function, i.e., $r = r_{\text{perf}} - \bar{r}_{\text{perf}}$, where   we use  REINFORCE~\cite{williams1992simple} to update the policy network with
\begin{equation}
L_{\text{reinforce}} = -r \log p(c_j|\widetilde{\mathbf{e}}_j).
\label{eq:5}
\end{equation}

\begin{algorithm}[t]
\caption{Learning procedure of ClusterP3S}
\label{alg:1}
\setlength{\intextsep}{0pt} 
\begin{algorithmic}[1]
\STATE \textbf{Input:} Training dataset $\mathcal{D}_{\text{train}}$, validation dataset  $\mathcal{D}_{\text{valid}}$, the search space of the preprocessing pipelines $\mathcal{P}$, a machine learning model $M$.
\STATE Learn feature embedding based on Eq.~(\ref{eq:3}).
\STATE Initialize the deep clustering network based on Eq.~(\ref{eq:4}).
\FOR{each upper-level iteration}
    \STATE Assign each feature to a cluster using the cluster network.
    \FOR{each lower-level iteration}
        \STATE Randomly sample a preprocessing pipeline within $\mathcal{P}$ based on the current clusters.
        \IF{the sampled pipeline achieves better performance}
            \STATE Store the sampled pipeline and mark it as the best
        \ENDIF
    \ENDFOR
    \STATE Update the clustering network using the best performance as the reward based on Eq.~(\ref{eq:5})
\ENDFOR
\end{algorithmic}
\end{algorithm}

\noindent \textbf{Summary.} We summarize the overall learning procedure in Algorithm~\ref{alg:1}. After initializing the feature embedding and the clustering network (lines 2 and 3), we jointly optimize the clustering network and search for the best preprocessing pipelines. In the outer loop, we generate a cluster assignment (line 5) and update the clustering network based on the feedback from the inner loop (line 12). In the inner loop, we use random search to search for the best preprocessing pipelines given a cluster assignment (lines 5 to 10). Finally, The best pre-possessing pipeline will be returned.

\subsection{Search Space Design}
\label{sec:method3}

This subsection introduces the search space of ClusterP3S. We include several standard preprocessing primitives, which can be grouped into three categories: (i) imputers that complete the missing values of the features, (ii) encoders that encode the features with transformation, and (iii) scalers that scale the feature values. We design a search space that allows each feature to be sequentially processed by an imputer, an encoder, and a scaler, whose order is fixed. However, a preprocessing pipeline can skip a specific primitive step. For example, it can only have an encoder and a scaler without an imputer. Our search space includes 11 primitives and 48 possible preprocessing pipelines. We provide more details of these primitives in \textbf{Appendix~\ref{apendix:1}}. Therefore, the overall search space is $48^D$, which grows exponentially with the number of features.

\section{Experiments}
We evaluate the performance of ClusterP3S on several benchmark datasets. Specifically, we aim to answer the following research questions. \textbf{RQ1:} Can ClusterP3S improve performance by enabling feature-wise preprocessing pipeline search (Section~\ref{sec:exp2})? \textbf{RQ2:} How effective of the proposed deep clustering network compared with vanilla clustering algorithm, such as K-Means (Section~\ref{sec:exp3})? \textbf{RQ3:} How efficient is the search of ClusterP3S (Section~\ref{sec:exp4})? \textbf{RQ4:} What is the impact of cluster numbers on ClusterP3S (Section~\ref{sec:exp5})? \textbf{RQ5:} How does the preprocessing pipeline discovered by ClusterP3S compare with the heuristic methods (Section~\ref{sec:exp6})?

\subsection{Experimental Setting}
\label{sec:exp1}
\noindent\textbf{Datasets.} For comprehensive comparison, we evaluate the performance of ClusterP3S on eight datasets with various scales and feature characteristics from the OpenMLcc-18 benchmark~\cite{bischl2017openml}. Specifically, these datasets differ in sample size and feature characteristics (e.g., missing value ratios and the number of numerical and categorical features). Table~\ref{tbl:dataset_stats} summarizes the data statistics. More details are provided in \textbf{Appendix~\ref{apendix:2}}.

% \textcolor{blue}{Maybe we can remove the following hyperlinks to the Appendix?} Datasets used from OpenMLcc-18 benchmark. 
% sick https://www.openml.org/d/38

% credit-a https://www.openml.org/d/29

% pca-4 https://www.openml.org/d/1049

% ilpd https://www.openml.org/d/1480

% car https://www.openml.org/d/40975

% tic-tac-toe https://www.openml.org/d/50

% kr-vs-kp https://www.openml.org/d/3

% dresses https://www.openml.org/d/23381

\noindent\textbf{Comparison Methods.} Since there is no automated search algorithm designed for feature-wise preprocessing, we implement two feasible solutions for comparison including heuristic pre-processing pipeline (\textbf{HeuristicP3}) and random clustering based personalized pre-processing pipeline (\textbf{RandClusterP3}). HeuristicP3 is obtained by identifying the optimal pre-processing configurations via common practice or domain experts. RandClusterP3 is obtained by randomly clustering features into different groups. The details of the comparison methods are as follows.

\begin{itemize}
    \item \textbf{HeuristicP3} is built upon a handcrafted pipeline that takes a series of widely adopted preprocessing primitives. For the case of numerical values, an imputation method will be applied if there are missing values. Specifically, we will first fill the missing values using the mean value of observed features and then normalize each feature via a scaler primitive MaxABS\footnote{\url{https://github.com/automl/auto-sklearn/tree/master/autosklearn/pipeline/components/feature_preprocessing}}. For the non-numerical features, we first fill the missing values using the most frequent value and then adopt a one-hot encoding technique to transfer the data. Finally, we concatenate the resultant numerical features and the transformed non-numerical features as input for downstream machine learners.
    
    % an imputation method using the most frequent value is added if the data contains missing data, then the data is transformed by using OneHot encoder. Finally, the a machine learning learner is added that uses as input the concatenation of the data transformation steps.
    \item \textbf{RandClusterP3} is a variant of the proposed framework. It is obtained by replacing the deep clustering network adopted by ClusterP3S with a random algorithm, which randomly assigns each feature into different group clusters. After generating a set of random clusters, we apply the same cluster-wise preprocessing pipeline used in ClusterP3S for model evaluation. 
    % consists on a variation of the propose method that instead of generating clusters using a Deep Clustering network uses clusters that are randomly assigned. After a set of random clusters has been generated, we apply Cluster-wise Preprocessing Pipeline Search.
\end{itemize}

\begin{table}[t]
\centering
\caption{Dataset Statistics.}
\vspace{-8pt}
\label{tbl:dataset_stats}
\footnotesize
\setlength{\tabcolsep}{2.5pt}
\begin{tabular}[t]{l*{5}{c}}
\toprule
Dataset             & \# Features & \# Samples & \# Missing values & \# Numerical Features & \# Non-numerical Features\\ 
\midrule
38-sick             & 30 & 3772 & 6064 & 6 & 22 \\
29-credit-a         & 16 & 690 & 67 & 6 & 9 \\
1049-pc4            & 38 & 1458 & 0 & 37 & 0 \\
1480-ilpd           & 11 & 583 & 0 & 9 & 1 \\
23381-dresses-sales & 13 & 500 & 835 & 1 & 11 \\
3-kr-vs-kp          & 37 & 3196 & 0 & 0 & 36 \\
40975-car           & 7  & 1728 & 0 & 0 & 6 \\
50-tic-tac-toe      & 10 & 958 & 0 & 0 & 9 \\
\bottomrule
\end{tabular}
\vspace{-15pt}
\end{table}

\noindent\textbf{Learning Protocol.} Following classical evaluation protocols for supervised setting~\cite{bischl2017openml,gijsbers2019open,probst2019tunability}, we first use the whole dataset to conduct pipeline search, and then apply 10-fold cross-validation to evaluate search preprocessing pipeline under different machine learning learners. We consider five popular machine learning classifiers: RandomForest, GradientBoosting, SVC, AdaBoost and SGDC from Sklearn \cite{sklearn_api}. To avoid the randomness, we repeat the process twenty times and report the averaged accuracy results.

% Evaluation method:

% Use 8 OpenMLcc-18 datasets that were selected to cover diverse scenarios (different data types, all non-numerical, missing values).

% For every baseline and proposed method, the whole dataset was available for search and evaluate the performance using 10-cross-fold validation. Papers where 10-fold:
% \cite{bischl2017openml}, \cite{gijsbers2019open}, \cite{probst2019tunability}

% For every dataset we use a set of machine learning learners: RandomForestClassifier, GradientBoostingClassifier, SVC, AdaBoostClassifier, SGDClassifier

% We ran our method for 50 iterations, the random clusters until we were able to generate a valid solution.

\noindent\textbf{Implementation Details.} We implement our model based on Pytorch and search 50 iterations using Adam optimizer with a learning rate equal to 0.001. For the AutoEncoder network, we adopt six fully connected layers with hidden dimensions fixed as 128. For the deep clustering network, we adopt four fully connected layers with hidden dimensions set to 128. We set number of clusters $K$ to be 5 for all the experiments. We provide more details in \textbf{Appendix~\ref{apendix:3}}.

% There is one crucial hyperparameter (i.e., the number of clusters $K$) in our model and random clustering-based variant. 

% For random and out method we limited the number of clusters $k=5$ since there are some small datasets. An ablation study of the impact of k is later. More details are provided in \textbf{Appendix~\ref{apendix:3}}.

% Networks:

% AutoEncoder: 6 fully connected layers, input/ouput layer=n\_features, intermediate layers width=128 with Adam optimizer and learning rate=1e-3

% DeepClustering: 4 fully connected layers, input layer=n\_features, output layer = n\_clusters, intermediate layers width=128 with Adam optimizer and learning rate=1e-3

\subsection{Comparison with Baselines}
\label{sec:exp2}
We start by comparing the performance of ClusterP3S with other baseline methods (\textbf{RQ1}). It is worth noting that the naive feature-wise random search solution is not involved for comparison. This is because it takes an extremely long time to scan the pre-processing pipeline candidates if the feature space is relatively large or the performance is inferior using the identified feature-wise pipelines. Table~\ref{tbl:fwt_acc_improve} reports the result over 8 datasets in terms of accuracy. We made several observations. \textbf{First}, the proposed ClusterP3S performs better than two baseline methods 
across all datasets. Specifically, ClusterP3S improves 3.5\% and 2.0\% over HeuristicP3 and RandClusterP3 on average, respectively. \textbf{Second}, although RandClusterP3 achieves comparable or even better better results than HeuristicP3 in most cases, it underperforms our method in all scenarios. The possible reason is that different features may favor various pre-processing pipelines, and our method can effectively identify specific pipelines for various features via reinforcement learning.   

Besides, we dig into the learning process of our method and the two competitors, shown in the left panel of Figure~\ref{fig:self}. The performances of our method and RandClusterP3 first increase with more iterations and then stay stable. ClusterP3S consistently outperforms RandClusterP3 during the learning process by a wide margin. We attribute it to our policy-based deep clustering network. 
% Mention that (naive solution) feature-wise random search either takes very long time and if a result is found, the performance is going to be poor, for that reason these results are not presented on the table.

% The reported results are the performance average across different learners (describe in eval protocol). overall the mean and std becomes lower with out method since the preprocessing for the clusters can be different transformations that are not limited by data-types as other methods.

% \begin{figure}[!ht]
% \centering
% \includegraphics[width=0.5\textwidth]{figures/performance.png}
% \caption{Performance comparison}
% \label{fig:performance}
% \end{figure}

% \begin{figure}[t!]
% \centering
% \includegraphics[width=0.8\textwidth]{figures/acc_improvement.png}
% \caption{Accuracy improvement on different machine learning learners focuses on heterogeneous data from the OpenMl-CC18 Benchmark with a 500 iteration search. The proposed method has more impact on the learners if the data contains different data types such as integer, boolean values. The performance improvement on machine learning learners that are tree-based is minimal since tree methods are already dealing with categorical data better.}
% \label{fig:fwt_acc_improvement}
% \end{figure}

\begin{table}[t]
\centering
\caption{Accuracy performance of all methods over eight OpenML-CC18 Benchmark datasets. The accuracy values are averaged based on five classical machine learning algorithms: RandomForest, GradientBoosting, SVC, AdaBoost, and SGD. 
% Comparison between the average accuracy of standard, random clustered and Clustered Feature-wise pipelines on OpenML-CC18 Benchmark datasets. CP3S shows a promising increase of performance for standard machine learners. \textcolor{red}{I have a concern: What is the base machine learning classifier used in this table? } It is the average across multiple ML learners RandomForestClassifier, GradientBoostingClassifier, SVC, AdaBoostClassifier, SGDClassifier
}
\label{tbl:fwt_acc_improve}
\footnotesize
\begin{tabular}[t]{lccc}
\toprule
Dataset             & HeuristicP3 & RandClusterP3 & ClusterP3S\\ 
\midrule
38-sick             & 96.63 $\pm$ 1.69 & 97.29 $\pm$ 1.36 & 97.70 $\pm$ 1.14\\
29-credit-a         & 84.28 $\pm$ 1.94 & 82.55 $\pm$ 4.76 & 85.07 $\pm$ 1.11\\
1049-pc4            & 88.69 $\pm$ 1.40 & 90.06 $\pm$ 0.67 & 91.08 $\pm$ 0.38\\
1480-ilpd           & 70.77 $\pm$ 2.07 & 70.60 $\pm$ 0.71 & 71.35 $\pm$ 1.09\\
23381-dresses-sales  & 58.04 $\pm$ 3.14 & 59.44 $\pm$ 1.61 & 62.20 $\pm$ 1.60\\
3-kr-vs-kp          & 94.77 $\pm$ 0.65 & 94.88 $\pm$ 1.18 & 95.18 $\pm$ 1.24\\
40975-car           & 86.42 $\pm$ 4.16 & 84.40 $\pm$ 3.73 & 86.70 $\pm$ 3.45\\
50-tic-tac-toe      & 77.52 $\pm$ 10.31 & 87.70 $\pm$ 7.23 & 91.00 $\pm$ 7.46\\
\midrule
Average             & 82.14 $\pm$ 3.17 & 83.36$\pm$ 2.66 & 85.03 $\pm$ 2.19\\

\bottomrule
\end{tabular}
\end{table}

% Learning "curve of hour method"

\subsection{Ablation Study}
\label{sec:exp3}
\begin{figure*}[t]
  \centering
  \begin{subfigure}[b]{0.33\textwidth}
    \centering
    \includegraphics[width=5.4cm]{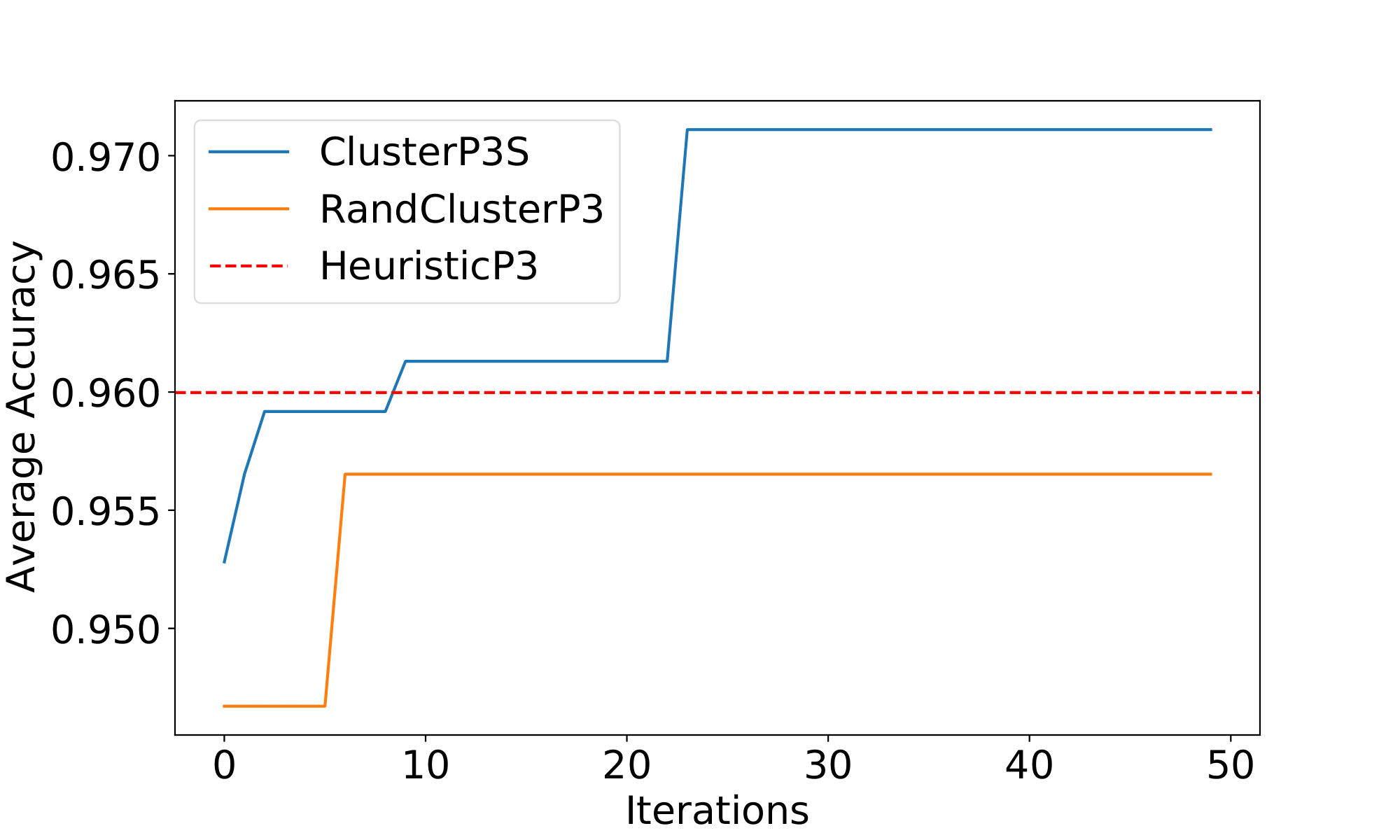}
    % \label{fig-ablation}
    % \caption{ClusterP3S vs. Kmeans.}
  \end{subfigure}%
  \begin{subfigure}[b]{0.33\textwidth}
    \centering
    \includegraphics[width=5.4cm]{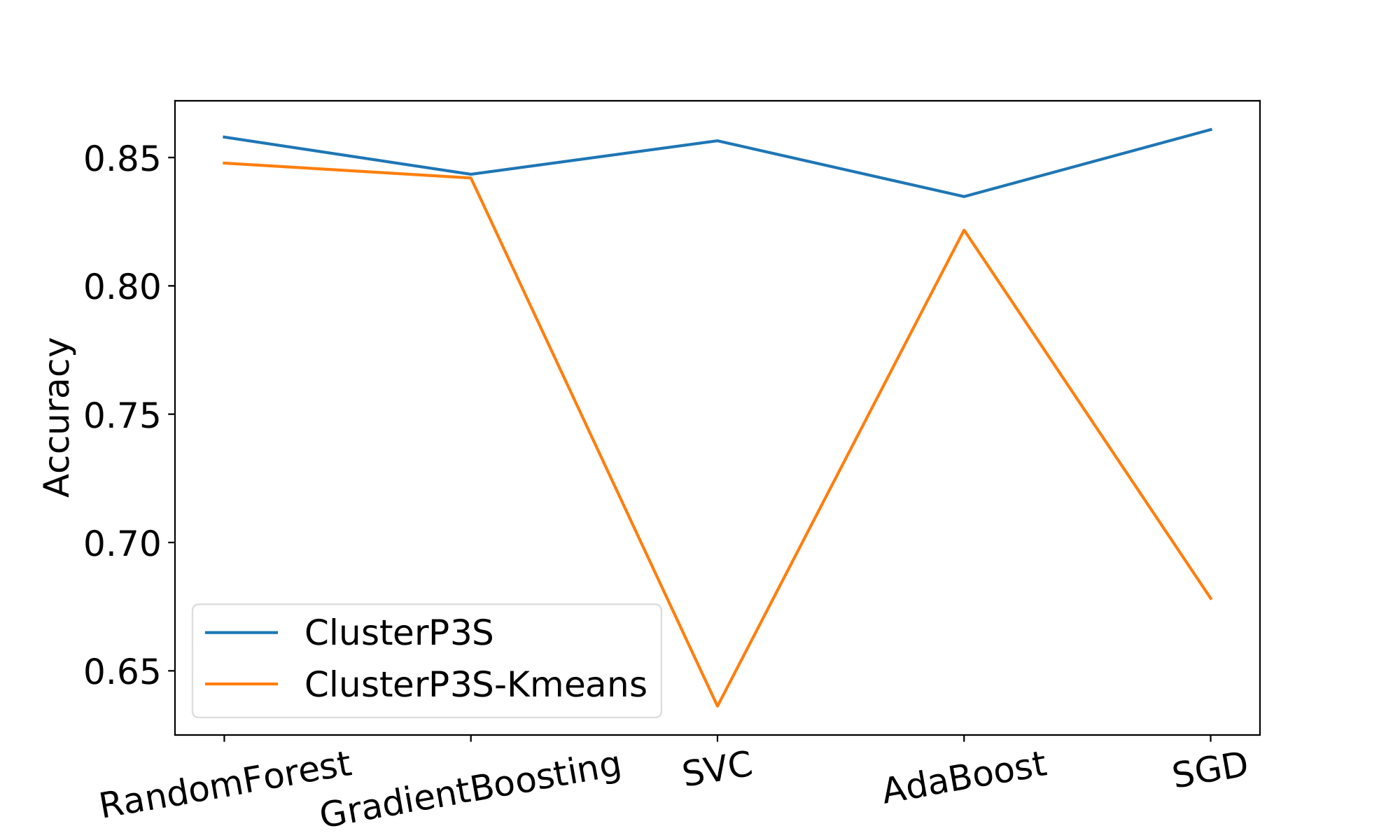}
    % \caption{Training iterations vs. model performance.}
  \end{subfigure}%
  \begin{subfigure}[b]{0.33\textwidth}
    \centering
    \includegraphics[width=5.4cm]{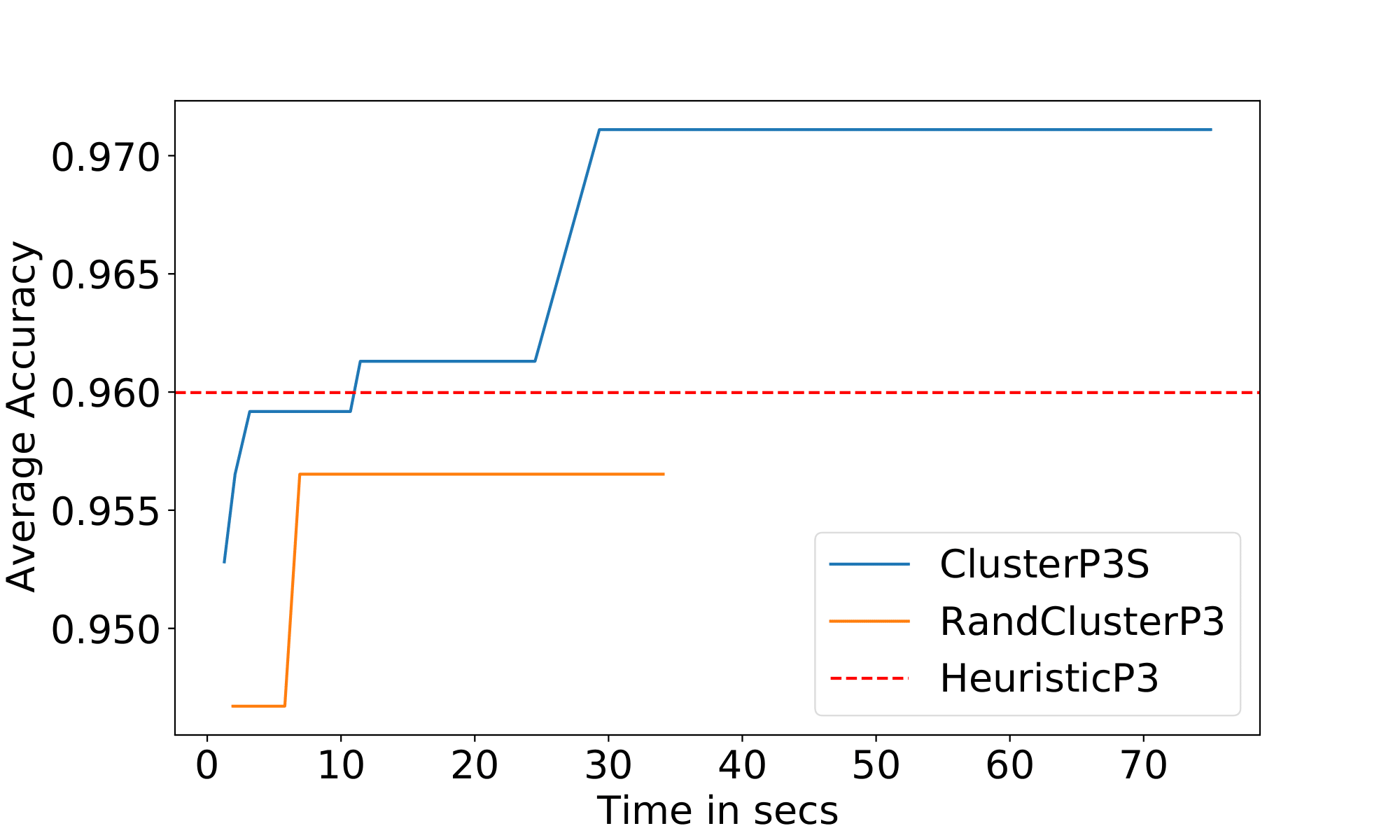}
  \end{subfigure}%
  \caption{ClusterP3S analysis. \textbf{Left:} The impact of learning iterations on ClusterP3S and baseline methods. \textbf{Middle:} The performance of ClusterP3S vs. ClusterP3S-Kmeans across five machine learning classifiers.  \textbf{Right:} Efficiency analysis of ClusterP3S.}
  \label{fig:self}
\end{figure*}

To answer question \textbf{RQ2}, we compare our model with a Kmeans based variant (ClusterP3S-Kmeans) across a set of machine learning classifiers in middle panel of Figure~\ref{fig:self}. ClusterP3S-Kmeans is obtained by replacing our policy-based deep clustering network with the popular K-Means algorithm. Our model performs consistently better than K-Means based variant across five different classifiers. Specifically, ClusterP3S significantly outperforms the K-Means variant on SVC and SGD base classifiers. The possible reason is that RandomForest, Gradient Boosting and AdaBoost are based on decision tree, which can capture discrete values better than other numerical methods. This result demonstrates the effectiveness of the proposed policy-based deep clustering network. 

% \begin{figure}[t]
% \centering
% \includegraphics[width=0.8\textwidth]{figures/ablation.pdf}
% \caption{Comparison of PC3S with KMeans using different ML learners and Datasets}
% \label{fig:ablation}
% \end{figure}

% \begin{figure}[t]
% \centering
% \includegraphics[width=0.8\textwidth]{figures/efficiency_acc_iterations.pdf}
% \caption{Improvement performance over iterations}
% \label{fig:improve_acc_iter}
% \end{figure}

% \begin{figure}[t]
% \centering
% \includegraphics[width=0.8\textwidth]{figures/efficiency_acc_times.pdf}
% \caption{Improvement performance over time}
% \label{fig:improve_acc_time}
% \end{figure}

\subsection{Efficiency Analysis}
\label{sec:exp4}
To investigate the efficiency of ClusterP3S (\textbf{RQ3}), following the setting in Section~\ref{sec:exp3}, we plot the averaged accuracy results of all methods with respect to the running time (in second) in Figure~\ref{fig:self} (the right panel). The x-axis denotes the wall-clock time in seconds. In general, our method achieves significantly better performance than the two baselines after only 15 seconds of search.
%by taking a long time slightly. For example, our proposed method can already achieve satisfactory results after 15 seconds.  

% Random clustering

% No clustering

% K-means

% remove Auto Encoder

\subsection{Hyperparameter Study}
\label{sec:exp5}
To analyze the impact of the number of clusters $K$ (\textbf{RQ4}), we vary $K$ from the set $\{2, 5, 10, 20\}$ and report the results on all datasets based on SVC classifier in Table~\ref{tbl:k_hyperparameter}. The results show that our method is not sensitive to the number of clusters when $K \ge 5$. A possible reason is that the deep clustering network can automatically learn the number of clusters even with a too large $K$.

% K, the number of layers.. 

% we apply or method to the same conditions reported on the evaluation protocol and we changed $k$ so we can study the effect on our method.

% Our method seems to adapt to the different $k$ 

\begin{table}[t]
\centering
\caption{The performance of ClusterP3S \textit{w.r.t.} $K$ based on SVC classifier. "-" means not applicable since these datasets have fewer number of features than $K$.}
\label{tbl:k_hyperparameter}
\vspace{-8pt}
\footnotesize
\begin{tabular}[t]{l|*{4}{c}}
\toprule
Dataset             & $K=2$ & $K=5$ & $K=10$ & $K=20$\\ 
\midrule

38-sick               & 97.61 & 96.81 & 96.81 & 97.29 \\
29-credit-a           & 84.78 & 85.65 & 85.65 & - \\
1049-pc4              & 90.73 & 90.80 & 91.01 & 90.80 \\
1480-ilpd             & 71.36 & 72.22 & 71.35 & - \\
23381-dresses-sales   & 58.0 & 62.2 & 62.4 & - \\
3-kr-vs-kp            & 93.89 & 94.24 & 94.49 & 94.30 \\
40975-car             & 89.75 & 89.75 & - & - \\
50-tic-tac-toe        & 93.21 & 93.21 & - & - \\

\bottomrule
\end{tabular}
\vspace{-15pt}
\end{table}

\subsection{Case Study}
\label{sec:exp6}
\begin{figure}[t]
\centering
\includegraphics[width=0.9\textwidth]{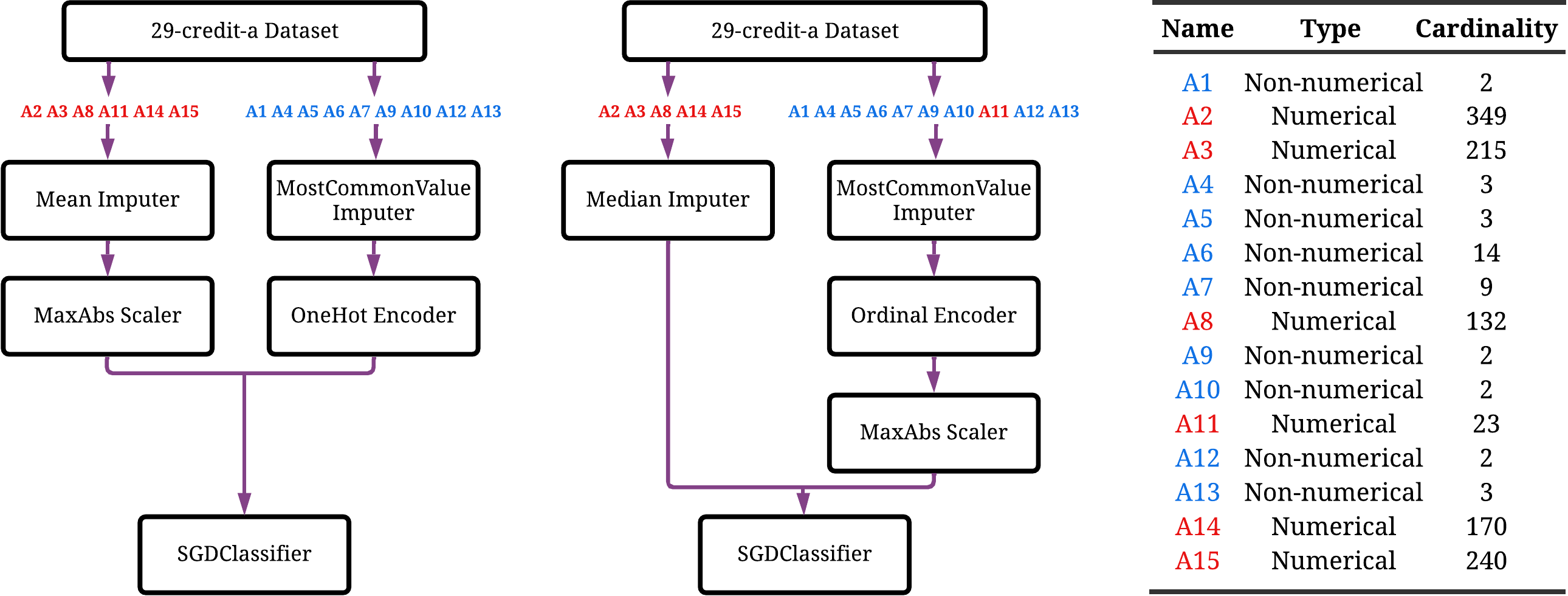}
\caption{Comparison of the heuristic pipeline (\textbf{left}) and the pipeline discovered by ClusterP3S (\textbf{middle}) on the 29-credit-a dataset. The table shows the characteristics of the features (\textbf{right}), where numerical features are marked in red, and non-numerical features are marked in blue. We can observe that feature A11 (which has relatively low cardinality) is grouped with non-numerical features even if it is essentially a numerical feature. }
\label{fig:diff_pipelines}
\end{figure}

To answer \textbf{RQ5}, we conduct a case study on the 29-credit-a dataset by comparing the heuristic pipeline and the one discovered by ClusterP3S, shown in Figure~\ref{fig:diff_pipelines}. The heuristic pipeline adopts two commonly used preprocessing pipelines for numerical and non-numerical features, respectively. Whereas, ClusterP3S groups feature A11 with non-numerical features. A possible explanation is that A11 has a low cardinality such that treating it like a non-numerical feature leads to better performance. We can also observe that the pipeline discovered by ClusterP3S does not quite align with our intuitions. For example, it is unclear why ordinal encoder is used. Interestingly, ClusterP3S can achieve better performance with this pipeline. This suggests the necessity of using AutoML, which could discover novel preprocessing pipelines that cannot be easily identified by humans.

%  Different pipelines left heuristic - right our method -the table on the right is about the feature cardinality

\section{Related Work}
\textbf{AutoML.} AutoML systems have achieved remarkable success in various ML design tasks, such as hyperparameter tuning~\cite{feurer2019hyperparameter,liaw2018tune}, algorithm selection~\cite{lindauer2015autofolio,mohr2021towards}, neural architecture search~\cite{klein2017fast,zimmer2021auto,zoph2016neural,wang2022auto,li2021automated}, meta-learning~\cite{finn2017model,vanschoren2018meta,behl2019alpha}, and pipeline search~\cite{thornton2013auto,feurer2020auto,heffetz2020deepline,drori2021alphad3m,yang2020automl,rakotoarison2019automated,kishimoto2021bandit,zha2021autovideo,milutinovic2020evaluation,lai2021tods,li2020pyodds,li2021autood,li2019pyodds,lai2021revisiting}. In this work, we study pipeline search and focus on pre-processing pipelines, which complements the existing efforts. We propose to enable feature-wise preprocessing pipeline search and devise a cluster-based algorithm to efficiently explore the search space.

\noindent\textbf{Reinforcement Learning.} Reinforcement learning has shown strong performance in many reward-driven tasks~\cite{mnih2013playing,zha2021douzero,schulman2017proximal,lillicrap2015continuous,silver2016mastering,silver2017mastering,jumper2021highly,zha2019experience,zha2021rank,lai2020dual,zha2021simplifying,zha2020meta,zha2021rlcard,zha2022autoshard,zha2022dreamshard,zha2022towards,zha2019rlcard,lai2020policy,dong2023active}. It has also been applied to AutoML search~\cite{zoph2016neural}. More recently, automated reinforcement learning has been explored~\cite{parker2022automated}. In this work, we reduce the search space with a novel deep clustering network, which is trained using reinforcement based on the reward obtained from the performance.

\section{Conclusions}
In this work, we study whether enabling feature-wise preprocessing pipeline search can improve performance. To tackle the large search space, we propose ClusterP3S, which jointly learns the clusters and search for the optimal pipelines with a deep clustering network trained by reinforcement learning. Experiments on benchmark datasets show that enabling feature-wise preprocessing pipeline search can significantly improve performance. We hope this insight and the idea of learning feature clusters can motivate future AutoML system designs.

\section{Limitations and Broader Impact Statement}

Preprocessing is a key step in building ML pipelines. Our work advances the existing algorithms by enabling feature-wise preprocessing pipeline search. Our efforts will pave the way towards generic, robust, and efficient AutoML systems, which will broadly benefit various ML-driven applications, such as recommender systems, healthcare, fraud detection, traffic prediction, etc. There is no negative societal impact to the best of our knowledge. Nevertheless, our research is limited in that we only search preprocessing pipelines. Combining ClusterP3S with model selection and hyperparameter tuning could lead to better performance, which we will investigate in the future. As fairness~\cite{chuang2022mitigating} becomes increasingly important, especially in high-stakes application~\cite{wan2022processing,ding2023fairly}, another future direction is fairness-aware preprocessing pipeline search. Making the search process more interpretable~\cite{chuang2023efficient,wang2022accelerating} is another direction that we plan to investigate.

% The 9 pages allocated for the main paper must include a discussion of limitations 
% and a broader impact statement regarding the approach, datasets and applications 
% proposed/used in your paper. It should reflect on the environmental, ethical and 
% societal implications of your work. The statement should require at most one page.
%
% This section is included in the template as a default, but you can also place these
% discussions anywhere else in the main paper, e.g., in the introduction/future work.

% The Centre for the Governance of AI has written an excellent guide for writing
% good broader impact statements (for the NeurIPS conference) that may be a
% useful resource for AutoML-Conf authors:
%
% https://medium.com/@GovAI/a-guide-to-writing-the-neurips-impact-statement-4293b723f832

% print bibliography -- for bibtex / natbib, use:

% \bibliography{...}
\bibliographystyle{abbrv}
\bibliography{references}
% and for biber / biblatex, use:

% \printbibliography

% supplemental material -- everything hereafter will be suppressed during
% submission time if the hidesupplement option is provided!
\appendix

\newpage
\section{Search Space Details}
\label{apendix:1}

\begin{table}[t]
\centering
\caption{preprocessing pipeline search space. The 11 primitives can be grouped into three categories: imputers, encoders, and scalars. Each category has different available options and includes the None operation that skips the step. }
\label{tbl:search_space}
\begin{tabular}[t]{ccc}
\toprule
Imputers          & Encoders & Scalers  \\
\midrule
Median            & Ordinal  & MinMax   \\
MostFrequentValue & OneHot  & Standard \\
Mean              & None     & MaxAbs   \\
None              & -        & None     \\
\bottomrule
\end{tabular}
\end{table}

The search space is summarized in Table~\ref{tbl:search_space}. The search space consist of imputers, encoders, and scalers. We will describe the included primitives under each of theses categories. Note that there is a None primitive under each category; it will do nothing and simply skip the primitive step. The details of the three imputers are as follows, we use the same imputation primitive SimpleImputer \footnote{\url{https://scikit-learn.org/stable/modules/generated/sklearn.impute.SimpleImputer.html}} and use the different strategies: 

\begin{itemize}
    \item \textbf{Median}: Fill the missing values using the median of each column.
    \item \textbf{MostFrequentValue}: Fill the missing using the most frequent value of each column
    \item \textbf{Mean}: Fill the missing values using the mean of each column.
\end{itemize}

The details of the two encoders are as follows.

\begin{itemize}
    \item \textbf{Ordinal}\footnote{\url{https://scikit-learn.org/stable/modules/generated/sklearn.preprocessing.OrdinalEncoder.html}}: 
    Encode categorical features as an integer array that are converted to ordinal integers.
    \item \textbf{OneHot}\footnote{\url{https://scikit-learn.org/stable/modules/generated/sklearn.preprocessing.OneHotEncoder.html}}: 
    Encode features as a one-hot numeric array that created a binary column for each category.
\end{itemize}

The details of the three scalers are as follows.

\begin{itemize}
    \item \textbf{MinMax}\footnote{\url{https://scikit-learn.org/stable/modules/generated/sklearn.preprocessing.MinMaxScaler.html}}: 
    Scaling each feature to a given range, in our case the default (0, 1).
    \item \textbf{Standard}\footnote{\url{https://scikit-learn.org/stable/modules/generated/sklearn.preprocessing.StandardScaler.html}}: 
    Standardize features by removing the mean and scaling to unit variance.
    \item \textbf{MaxAbs}\footnote{\url{https://scikit-learn.org/stable/modules/generated/sklearn.preprocessing.MaxAbsScaler.html}}: 
    Scale each feature by its maximum absolute value.
\end{itemize}

\section{Dataset Details }
\label{apendix:2}
The experiments are conducted on eight classification datasets with various characteristics from UCI Machine Learning Repository \cite{asuncion2007uci} via the OpenMLcc-18 benchmark \cite{bischl2017openml}. All the datasets are publicly available. We provide the detailed descriptions of these datasets below.

\pagebreak

\begin{itemize}

    \item \textbf{38-sick}\footnote{\url{https://www.openml.org/d/38}} This dataset contains 30 features, of which one is discarded since it is a feature column with only Null values. It contains 3772 samples, 6064 missing values, six numerical features, and 22 non-numerical features. The dataset aims to classify the samples into two categories, sick and negative.
    
    \item \textbf{29-credit-a}\footnote{\url{https://www.openml.org/d/29}}: This dataset contains 16 features and 690 samples. It has 67 missing values, six numerical features, and nine non-numerical features. The dataset is related to credit card applications and aims to classify the samples into two categories - and +.
    
    \item \textbf{1049-pc4}\footnote{\url{https://www.openml.org/d/1049}}: This dataset contains 38 features and 1458 samples. It does not contain any missing value, and all the features are numerical. The dataset aims to characterize code features associated with software quality and classify the sample into True and false categories.
    
    \item \textbf{1480-ilpd}\footnote{\url{https://www.openml.org/d/1480}}: This dataset contains 11 features and 583 samples. It does not contain any missing value and has nine numerical and one non-numerical feature. The dataset is a record of liver and non-liver patients and aims to classify whether or not a patient is a liver patient with class categories 1 and 2. 
    
    \item \textbf{23381-dresses-sales}\footnote{\url{https://www.openml.org/d/23381}}: This dataset contains 13 features and 500 samples. It has 835 missing values, one numerical feature, and 11 non-numerical features. The dataset contains attributes of dresses and their recommendations according to their sales with categories 1 and 2.
    
    \item \textbf{3-kr-vs-kp}\footnote{\url{https://www.openml.org/d/3}}: This dataset contains 37 features and 3196 samples. It has 835 missing values, one numerical feature, and 11 non-numerical features. This dataset aims to determine whether or not a chess configuration will result in a win or not.
    
    \item \textbf{40975-car}\footnote{\url{https://www.openml.org/d/40975}}: This dataset contains 7 features and 1728 samples. It does not contain any missing value and it has six non-numerical features. The dataset aims to classify whether or not a car is acceptable (acc) or unacceptable (unacc).
    
    \item \textbf{50-tic-tac-toe}\footnote{\url{https://www.openml.org/d/50}}: This datset contains 10 features and 958 samples. It does not contain any missing value and it has nine non-numerical features. The dataset aims to determine if the sample is a winner on a tic-tac-toe match, by using the categories positive and negative.

\end{itemize}

\section{Implementation Details}
\label{apendix:3}

Implementation can be found at \url{https://anonymous.4open.science/r/ClusterP3S-2B6B}

\subsection{Neural Architecture and Hyperparameter Configuration}

We list the neural architecture and hyperparameter configurations of ClusterP3S below.
\begin{itemize}
    \item \textbf{AutoEncoder}: Our Autoencoder implementation consists on six fully connected layers with hidden dimensions fixed as 128. The input layer dimension depends on the feature embedding using the term frequency of each column feature.
    \item \textbf{Clustering network}: Our deep clustering network implementation consists on four fully connected layers with hidden dimensions set to 128. We define a hyperparameter $K$ (maximum number of clusters) that is use as the output layer dimension.
    
    \item \textbf{Optimizer}: Adam optimizer with a learning rate equal to 0.001.
\end{itemize}

\subsection{Hardware}
For our experiments, we limited our hardware resources to 4 CPU cores (Intel(R) Xeon(R) CPU E5-2623) and 16 GB of Memory.

\end{document}